# A New COLD Feature based Handwriting Analysis for Enthnicity/Nationality Identification

[1]Sauradip Nag, [2]Palaiahnakote Shivakumara, [3]Wu Yirui, [4]Umapada Pal, and [5]Tong Lu

[1]Kalyani Government Engineering College, Kalyani, Kolkata, India

[2]Faculty of Computer Science and Information Technology, University of Malaya, Kuala Lumpur, Malaysia

[3]College of Computer and Information, Hohai University, Nanjing, China

[4]Computer Vision and Pattern Recognition Unit, Indian Statistical Institute, Kolkata, India.

[5]National Key Lab for Novel Software Technology, Nanjing University, Nanjing, China.

Email: sauradipnag95@gmail.com, shiva@um.edu.my, wuyirui@hhu.edu.cn, umapada@isical.ac.in, lutong@nju.edu.cn,

*Abstract*—Identifying crime for forensic investigating teams when crimes involve people of different nationals is challenging. This paper proposes a new method for ethnicity (nationality) identification based on Cloud of Line Distribution (COLD) features of handwriting components. The proposed method, at first, explores tangent angle for the contour pixels in each row and the mean of intensity values of each row in an image for segmenting text lines. For segmented text lines, we use tangent angle and direction of base lines to remove rule lines in the image. We use polygonal approximation for finding dominant points for contours of edge components. Then the proposed method connects the nearest dominant points of every dominant point, which results in line segments of dominant point pairs. For each line segment, the proposed method estimates angle and length, which gives a point in polar domain. For all the line segments, the proposed method generates dense points in polar domain, which results in COLD distribution. As character component shapes change, according to nationals, the shape of the distribution changes. This observation is extracted based on distance from pixels of distribution to Principal Axis of the distribution. Then the features are subjected to an SVM classifier for identifying nationals. Experiments are conducted on a complex dataset, which show the proposed method is effective and outperforms the existing method.

## I. Introduction

Traits, namely, gender, nationality, age, height, gait, etc., are popular in the field of biometric applications, such as face and iris recognition [1, 2]. This is because traits prediction helps biometric methods to improve their performances by reducing the complexity of the problem [3, 4]. In addition, traits prediction plays a vital role for forensic applications and security by helping in identifying suspicious behaviors [1]. However, it is noted from biometric based methods that when the input images are posed to an open environment, the methods lose accuracy. This is due to inherent limitations of biometric based methods, such as sensitivity to non-uniform illumination effect, occlusion, degradation, and environment influences. Besides, the methods are said to be computationally expensive as it involves high-level image processing tasks. As a result, in order to help forensic applications and investigation teams, handwriting analysis has received a special attention of the researchers, which has now reached beyond traditional boundaries such as emotions, nationality, gender, age and other traits prediction [4, 5]. However, due to large variations in handwriting, ink, pen, paper, script, age, gender, and individual difference, it is not so easy to identify traits based on handwriting analysis accurately. Therefore, this work focuses on nationality and ethnicity identification as it is useful for identifying crimes where different nationals are involved.

Despite the problem is challenging as mentioned above, one can expect differences in writing styles of different nationals. For example, Chinese usually prefer to write letters with more straight than cursive. This is valid because of the nature of their national language, where alphabets and text usually are formed with the combination of straight strokes. In case of people originating from India and Bangladesh, we can expect more cursive than straightness compared to Chinese because most of Indian and Bangladesh scripts are cursive in nature. With this notion, one can confirm that English writing changes from one national to another. This is the main basis for proposing the method in this work. It is evident from the sample images of each nation shown in Fig. 1, where it can be seen that each nation has a different writing style. At the same time, since all the citizens of respective nations follow their own scripts, we can expect English writing by different persons of the same nation share the common properties.

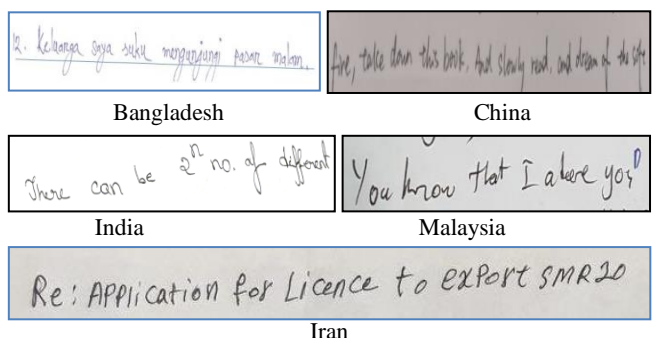

| Bangladesh | China |
| India | Malaysia |

Iran

Fig. 1. Examples of English handwriting of different countries

Hardly, we find a method that addresses the issue of nationality and ethnicity identification using handwriting analysis. Maadeed and Hassaine [6] proposed automatic prediction of age, gender, and nationality in offline handwriting. The method extracts direction, curvature, tortuosity, chain code and edge based directional features for character components. They believe that handwriting documents written by different nationals exhibit the unique difference. Based on this observation, the method extracts the above-mentioned features, which are then passed to a Random Forest classifier for the final classification. This work considers eight nationals for identification. However, the method reports poor results for nationality identification because the proposed features are inadequate to handle intra class variations.

It is noted from the above review that the problem of nationality/ethnicity identification is at infancy stage and hence there is a huge gap from 2014 to 2018. This fact motivated us to propose a new method for nationality identification in this work. The scope of this work is limited to five nationals, namely, Bangladesh, China, India, Malaysia and Iran. The reason to choose these nationals is that we felt English documents written by these nationals are challenging because they share the common ASIAN culture .In addition, we can see most of the above-mentioned countries have multi-religion, multi-lingual, etc. Therefore, this paper presents a method based on the fact that when nations have different scripts for writing, English handwriting by different nationals must have the unique difference in terms of writing style, spacing between characters and words, stroke pattern, etc. To extract such observation, motivated by the method [7] where Cloud of Line Distribution (COLD) has been used for writer identification, we explore the same with new feature extraction for nationality identification in this work. As content changes in images according to different nationals, the COLD distribution in polar coordinate domain changes. With this notion, we propose to explore shape of COLD distribution for extracting features such that the method identifies nationals, accurately.

## II. PROPOSED METHODOLOGY

This work considers English handwritten documents of five nationals, namely, Bangladesh, China, India, Malaysia and Iran for identification as mentioned in the previous section. Since the proposed work aims at identification of nationals at text line level, we segment text lines from handwritten documents based on horizontal projection profile analysis computed using tangent angle and the mean of intensity values of each row in the image. It can be expected that persons can use ruled or unruled page for writing. When a document image contains ruler, it poses a problem for extracting text components from text line images as it connects text components of lines as one big component. Therefore, the proposed work removes ruler from images based on tangent angle between character contours and ruler. As nationality changes, writing style also changes. It is valid because every national has the own unique national script and uses the same script for writing most time.

Motivated by the method [7], which explores COLD features for writer identification, we propose the same COLD for nationality identification through handwritten text lines analysis as it helps in studying character shapes of handwritten text lines. This gives different shapes of COLD distribution for different nationals, and gives almost the same distribution for writers of respective nations. In other words, the shape of distribution shares the common properties for text lines written by different writers of the same nation. To study shape of COLD distribution, we propose a new way of feature extraction by estimating distance from pixels to the principal axis of the distribution, which gives a feature matrix. The reason to choose principal axis is that it is independent of rotation, and is insensitive to pixels of the distribution because the principal axis is drawn based on the direction of majority pixels. That is if a distribution loses a few pixels, it does not affect for studying its shape through the principal axis. Furthermore, the feature matrix is subjected to an SVM classifier for nationality identification. The steps and flow of the proposed method can be seen in Fig. 1.

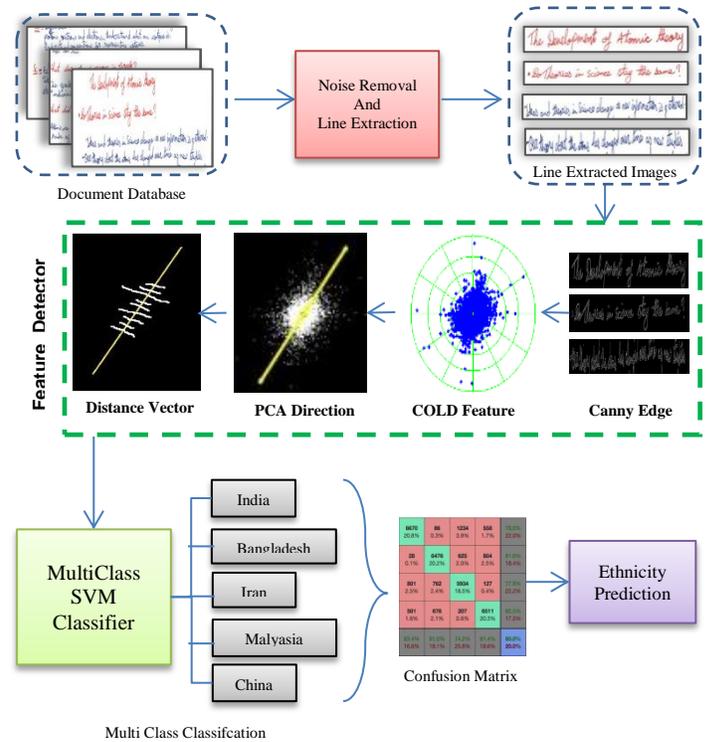

Fig 2 . Represents Proposed Workflow

### A. Preprocessing for Text Components Detection

Since the goal of the proposed works is to identify nationality at text line level, the proposed method segments text lines from handwritten documents based on horizontal projection profile(HPP) analysis. Motivated by the idea used by Koerich et al [9] we integrated Horizontal Projection Profile and Tangent Angle for each pixel in a row to eliminate baseline. The document image is first grayscaled and for each row , the mean of all the intensities $c_i(x_i, y_i)$ is calculated .

$$HPP(y) = \frac{\sum_{i=1}^{n} c_i(x_i, y_i)}{n} \qquad (1)$$

The method finds the mean of intensity values in a row and for each pixel in the row , we check if it is part of Canny Edge Component and compute the tangent of that pixel using the adjacent pixel in the Component using Equation(2). If the tangent angle is close to zero plus some threshold , then we can conclude that it is part of a baseline .

$$\emptyset = argtan\left(\frac{canny(y_i)-canny(y_{i+1})}{canny(x_i)-canny(x_{i+1})}\right) \quad (2)$$

It is true that when there is a space, the mean of intensity values is near to zero value. On the other hand, when there is a text, the mean gives a high value. At the same time, tangent angle for pixels in the middle row of a text area gives arbitrary angles, while for the pixels at the border, which are nearer to the baseline, they give almost zero angle. So the condition by which baselines are removed are given in Equation(3) below :

$$im(x,y) = \begin{cases} 0, & if\ HPP(y) > 200\ and\ \emptyset < 10° \\ c(x,y), & otherwise \end{cases} \quad (3)$$

The value of Horizontal Project Profile mean intensity and tangent angle $\emptyset$ is empirical and found after suitable experiments on skewed and unskewed document images. With the mean and tangent angle information, the proposed method performs horizontal projection profile analysis for segmenting text lines from handwritten documents.

For removing ruled lines from the segmented text line image, the proposed method finds the pixels that move parallel to tangent angle. It is valid because the pixels of a ruled line and the baseline have the same direction. The same observation is exploited for removing ruled lines in the image as shown in the sample results of Fig. 3, where in (a) we can see a ruled line and in (b) the ruled line is removed. This idea also works well for text line images that do not have ruled lines. In case of text line without ruled lines, the proposed method does not find pixels that move parallel to tangent angle. For the output of the pre-processing steps, the proposed method obtains Canny edge images for respective five nations handwritten images as shown in Fig 4. The edge components in the images are considered for generating COLD distribution, which will be discussed in the next section.

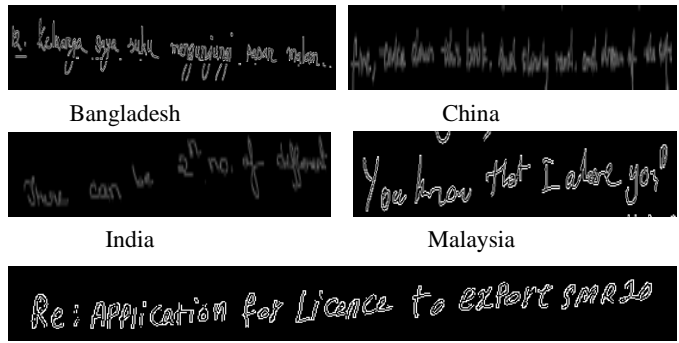

(a) Segmented text line from the ruled Bangladesh document

(b) Rule line is removed from the image in (a).
Fig. 3. Preprocessing step for feature extraction.

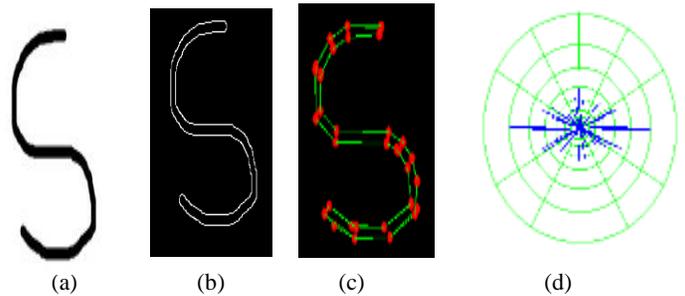

Bangladesh      China

India      Malaysia

Iran

Fig. 4. Canny edge images of handwritten line images of five nations.

## B. COLD Distribution for Feature Extraction

As discussed in the Introduction Section, the variations in handwriting of different nationals reflect in shapes of characters, while character shapes reflect in the form of strokes. To study the behavior of strokes in offline, it is necessary to understand curvy and straightness of strokes. Therefore, inspired by the method [7] where Cloud of Line Distribution (COLD) is proposed for writer identification, we propose the same for extracting variations in handwriting of different nations. For each edge component in the text line images as shown in Fig. 5(a) and Fig. 5(b), the proposed method finds dominant points for the contours using polygonal approximation algorithm [7] as shown in Fig. 5(c).

Let dominant points be over contour of edge components, $D = \{ P_i(x_i, y_i) | i = 1,2,3 .... N\}$ where $P$ denotes a dominant pixel, $(x_i, y_i)$ denotes $x$ and $y$ coordinates of the dominant pixel, and $N$ is the total number of dominant points. With these dominant points, the proposed method forms lines by joining nearest neighbor dominant points as shown in Fig. 5(c) where we can see dominant points pair, which results in line segments. To study the straightness and curviness of the line segments in polar coordinate domain, the proposed method estimates angle, $\theta$ for the line segments of the contour using x and y coordinates as defined in Equation (4). The length, $r$ of line segments is determined as defined in Equation (5).

$$\theta = \tan^{-1}\left(\frac{y_{i+1}-y_i}{x_{i+1}-x_i}\right) \quad (4)$$

$$r = abs\left(\sqrt{(y_{i+1}-y_i)^2 + (x_{i+1}-x_i)^2}\right) \quad (5)$$

Here $(x_i, y_i)$ and $(x_{i+1}, y_{i+1})$ denote the coordinates of a dominant pair. A line segment can be represented using $\theta$ and $r$ as a point $(\theta, r)$ in polar domain. We draw points for all the line segments in polar domain, which results in a distribution as shown in Fig. 5(d), where one can see COLD distribution for the line segments in Fig. 5(c).

(a)      (b)      (c)      (d)

Fig. 5. COLD distribution for handwriting components. (a) is a handwritten character, (b) is the Canny edge image of (a), (c) shows the dominant points for the contours, and (d) gives Cloud of Line Distribution (COLD) in polar coordinate.

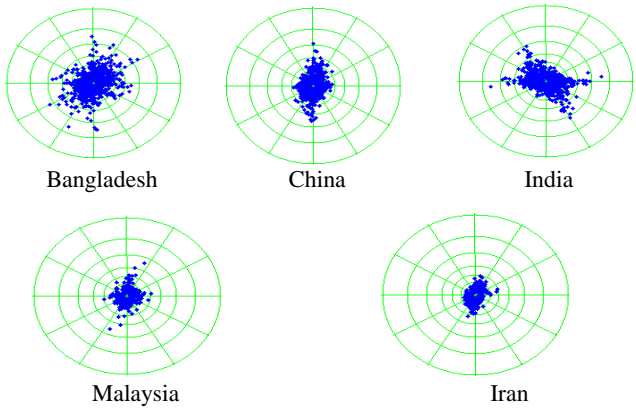

Fig. 6. COLD distribution of text lines of five nations.

Bangladesh  China  India
Malaysia  Iran

respectively . After this step these polar points are plotted in a plane of Radius 150 units . For each Line segment obtained from documents of a particular class , a COLD Plot is drawn and finally all the line segments of a particular Ethnicity is merged into one plot which reflects the variation of writers of a particular Ethnic group.

The shape of COLD changes according to the behavior of line segments of character components. For example, if a character is too cursive, in this case, one should expect more short line segments. When we have more short lines, the COLD distribution appears dense at the center origin. If line segments are too long, one can expect scattered COLD distribution. In other words, if character components have regular curvature, it is expected centralized COLD distribution with a particular angle. If character components have irregular curvature, it is expected scattered COLD distribution with arbitrary angle. It is evident from the COLD distribution of five nationals shown in Fig. 6, where one can see different distributions for all the five nationals. It is observed from Fig. 6 that since handwriting of Chinese nationals involves more vertical and horizontal strokes, the COLD distribution appars dense at the center region without scattered pixels. The same thing is true for handwriting of Malaysia and Iran with tiny differences. However, for Bangladesh and India, since they write many cursive characters that may involve regular and irregular curvature, the COLD distributions appear scattered with different orientations. These observations motivated us to extract features to study shapes of distributions for nationality identification.

As shown in Fig 7 , the Document Image written by a anonymous Person of a particular ethnicity is passed into the architecture pipeline . This image may contain Noise in the form of Ruled Lines over which handwriting is done . This Noise is removed using Horizontal Projection Profile as discussed in section II(A) . After this operation the individual Lines are segmented manually . After this step canny edge components are extracted individually from the segmented lines . These components are then passed into KeyPoint Extractor as shown in Fig 7 which uses Polygonal Approximation Method on the Canny Edge to reduce the number of Contour Points to bare minimum such that only the points which strongly define a particular character contour is only present. These Adjacent Keypoints are joined and passed into Transformer as shown in Figure 7. In this the distance between 2 adjacent keypoint of the character contour is considered and the coordinates of these two point is used to transform cartesian to polar coordinates using Eq 1 and 2

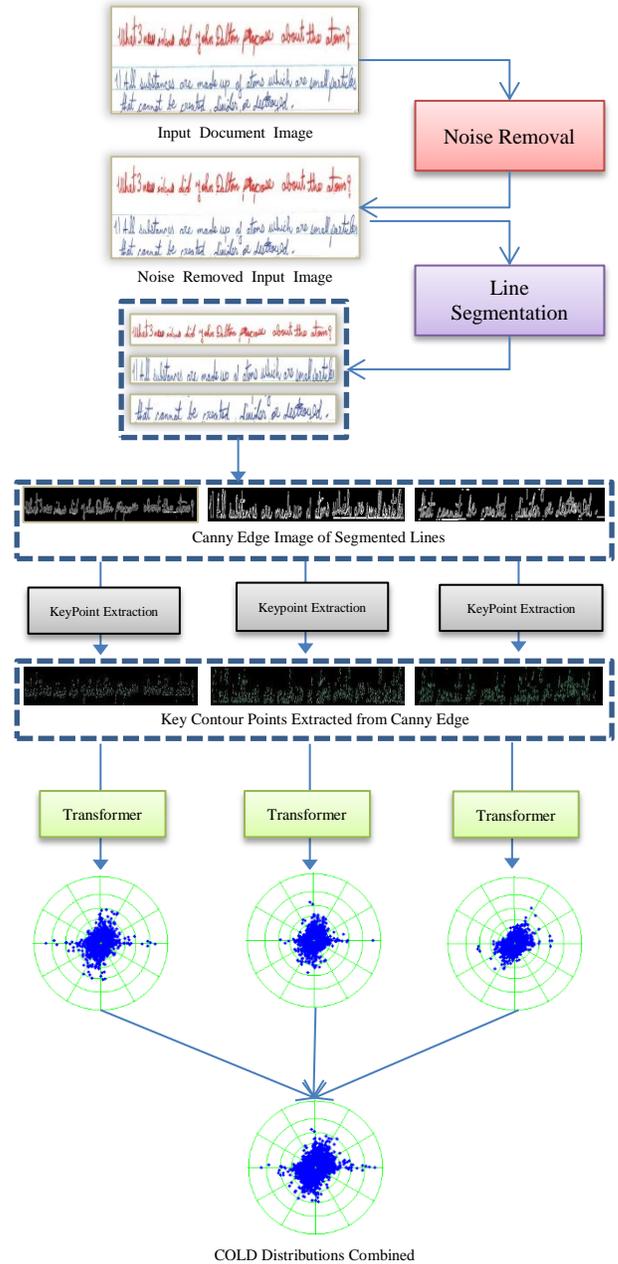

Fig. 7. Represents Flowchart of COLD Distribution Construction from Input

## C. Feature Extraction for Nationality Identification

For a COLD distribution, the proposed method maps all the pixels from polar domain to spatial domain, which results in a binary image as shown in the middle column in Fig. 8. This is to reduce the implementation difficulty. For studying shape of distribution, we propose to estimate principal axis with Principal Component Analysis (PCA), which gives a direction according to majority pixel directions and hence it is invariant to rotation and scaling. In addition, if a distribution loses a few pixels, it does not affect for drawing correct principal axis according to the shape of the distribution. The proposed method considers principal axis as a reference line and moves in perpendicular direction to every pixel of reference line in both left and right directions until it reaches white pixels of the distribution in both left and right direction. When it reaches white pixels in both directions, the proposed method calculates the distance from the left pixel to the reference, that is *LD*, and the right pixel to the reference, that is, *RD*. The absolute difference of *LD* and *RD* is considered as the feature of the pair of pixels. This process continues until it finds no pixel in any of left or right direction of reference while moving in the same direction. The whole process is repeated for all the pixels of the reference line. It is illustrated in the third column in Fig. 8, where we can see the reference line is marked by yellow color and lines are drawn for estimating the distance between the left and right side pixels for COLD distributions of five nations. Finding pair pixels and absolute difference of distance help us to withstand intra variations caused by different writers of the same nation. This process outputs a feature matrix for the input text line image.

Since we believe the proposed method extracts powerful features for nationality identification, we propose to use simple and well-known SVM classifier in this work rather than the recent learning tools such as deep learning.

For training and testing, we follow 10 fold cross validation procedure for identification. The size of the entire Dataset is 500 text line images . Hence we used 450 Randomly chosen text line Images for Training the MultiClass SVM  and 50 Randomly chosen text line image which is not part of Training Set is chosen for Testing the Trained SVM model. The Kernel used for SVM MultiClass Classification is Gaussian Kernel and the multiclass support is handled according to a one-vs-one scheme.The Kernel Cache Size was kept to 200MB during experiments .The Flag to set Scores to Posterior Probabilities was turned to true . The Hyperparameters are optimized using Bayessian Optimisation Technique

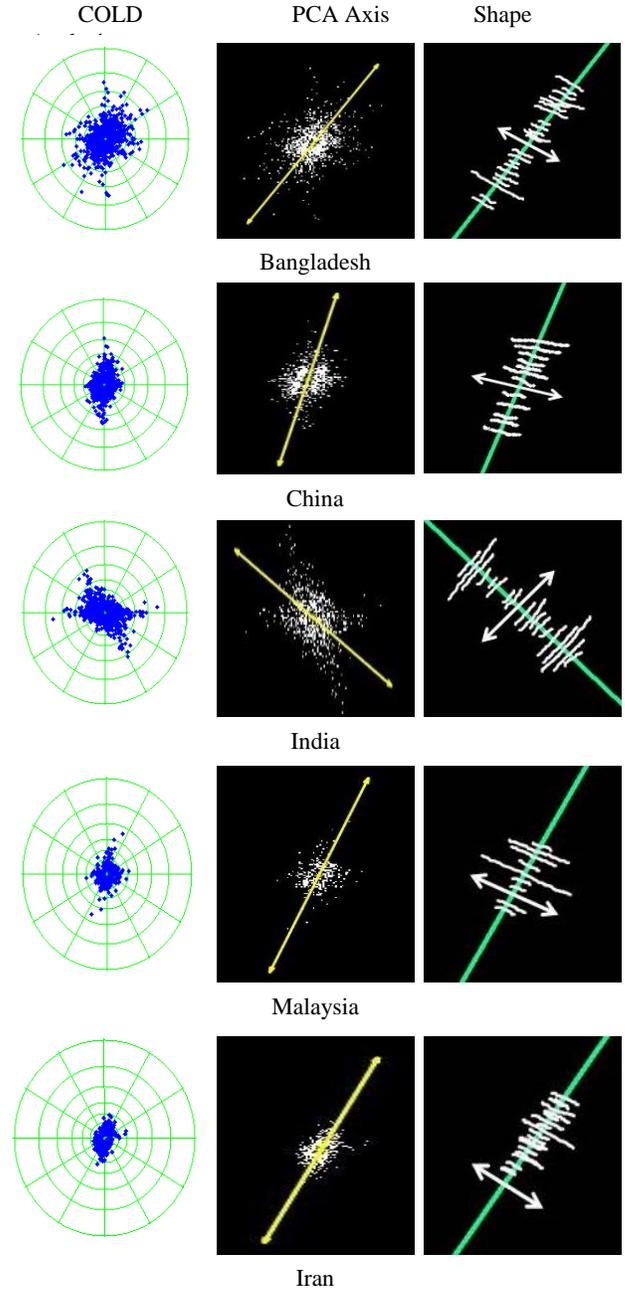

Fig. 8. Feature extraction for handwriting text lines of five nations.

## III. EXPERIMENTAL RESULTS

As per our knowledge, the dataset for nationals, which involves ASIAN countries, is not available publicly. We, thus, create our own dataset by collecting English handwriting images from different ASIAN countries. There is one dataset called QUWI [8], which is available but it considers countries which use Arabic as national language and does not involve any ASIAN countries, such as India and Malaysia. Our dataset consists of English handwritten line images of different aged persons from 20-25 years who are free to use any pen, ink, paper and topic for writing. The dataset comprises of 100 writers of a particular nationality who currently reside in their natives and this does not include handwritings of expatriates. Out of these data roughly 75 % of writers are Right Handed and rest 25% are Left Handed . The Gender of 90% of all collected data are Male and Rest 10% are female.Hence the dataset created for 5 classes is quite useful for multifaceted feature extraction for forensic works. As a result, the dataset is considered as complex data for validating the proposed method. We consider each nation as one class and 100 text line images per class are considered.These text line images are taken from 1 Half Page Handwritten Document with 1 text line per Half Page Document which are randomly cropped to maintain robustness. Therefore, for five nations, namely, Bangladesh, China, India, Malaysia and Iran, 500 text line images are considered for evaluating the proposed method.

For measuring the performance of the proposed method, we consider confusion matrix and Classification Rate (CR) as in the method [6] for nationality identification. To show the usefulness of the proposed method, we implement the method [6], which uses several geometric features based on contours to study the shape of character components and then identifies nationals for comparative study. However, the method is limited to nationals that consider Arabic as national language. In addition, the method reports very poor results for different features for nationality identification. This shows that the proposed geometrical features are not robust to intra and inter variations of handwritten text line images written by different nationals. The main reason to consider this method for comparative study here is that this is the state-of-the-art method for nationality identification. Though nationals are different, objective of the method is the same as the proposed method.

Sample successful handwritten text line images of five nations are shown in Fig. 9, where one can notice that our dataset includes low contrast text line images affected by blur, ruled and unruled pages. Quantitative results of the proposed and existing methods are reported in Table I and Table II, respectively, where it can be noticed that the proposed method outperforms the existing method in terms of classification rate. This is valid because the proposed features especially COLD distribution is robust to intra variations of different writers of the same nation, at the same time, it provides the unique shape for handwritten text line images of different nations. In addition, the proposed features are invariant to rotation, scaling and to some extent to distortion caused by low contrast, resolution and blur. On the other hand, the proposed features in the existing method are sensitive to pixels and contours. In other words, if a character or contour loses a few pixels, the method may not work well. Therefore, the existing method reports low results compared to the proposed method. Overall, the proposed method can be extended to other nations with more writers for each class because the features are robust and generic. This is the main advantage of the proposed method compared to the existing method.

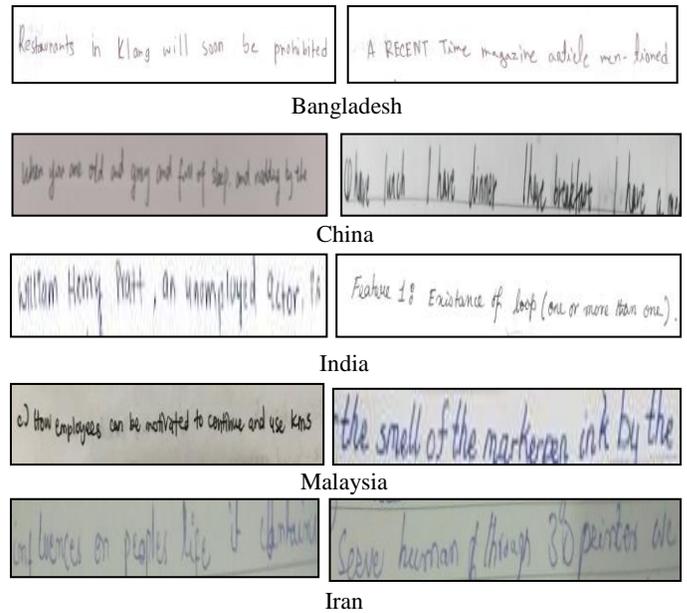

Fig. 9. Sample successful handwritten text line images of the proposed method for five ASIAN countries.

Table I. Confusion matrix and Classification Rate (CR) of the proposed method.

| Classes | Bangladesh | India | China | Iran | Malaysia |
|---|---|---|---|---|---|
| Bangladesh | **79%** | 14% | 0% | 4% | 3% |
| India | 15% | **73%** | 4% | 6% | 2% |
| China | 3% | 7% | **82%** | 3% | 5% |
| Iran | 7% | 13% | 3% | **66%** | 11% |
| Malaysia | 6% | 10% | 7% | 2% | **75%** |
| CR in (%) | 75 | | | | |

Table II. Confusion matrix and Classification Rate (CR) of the existing method [6].

| Classes | Bangladesh | India | China | Iran | Malaysia |
|---|---|---|---|---|---|
| Bangladesh | **41%** | 18% | 10% | 11% | 20% |
| India | 21% | **37%** | 15% | 13% | 14% |
| China | 12% | 20% | **38%** | 12% | 18% |
| Iran | 19% | 17% | 9% | **45%** | 10% |
| Malaysia | 17% | 10% | 23% | 18% | **32%** |
| CR in (%) | 38.6 | | | | |

IV. CONCLUSION AND FUTURE WORK

In this paper, we have proposed a new method for nationality identification. The proposed method does preprocessing for text line segmentation from handwritten document images and rule line elimination from the text line images using tangent angle and mean of intensity values. The proposed method explores the Cloud of Line Distribution (COLD) for studying shapes of character components in terms of straightness and curviness of strokes in polar domain. The unique shapes of distributions of different nations are studied based on new features given by principal axis and distance from pixel to the principal axis of each distribution. The feature matrix is subjected to an SVM classifier for nationality identification. In the future, we plan to extend the idea for identifying provinces in multi-lingual countries, such as India, Singapore and Malaysia, where we can see multiple scripts for different provinces.

ACKNOWLEDGEMENT